\pdfoutput=1
\documentclass[11pt]{article}
\usepackage[utf8]{inputenc}
\usepackage{graphicx}
\usepackage{amsmath}
\usepackage{booktabs}
\usepackage{multirow}
\usepackage{adjustbox}
\usepackage{makecell}
\usepackage{mdframed}

\usepackage[]{acl} 

\usepackage{times}
\usepackage{latexsym}

\usepackage[T1]{fontenc}
\usepackage[utf8]{inputenc}

\usepackage{microtype}

\usepackage{hyperref}
\usepackage{cleveref}
\usepackage{multirow}
\usepackage{graphicx}
\usepackage{pdfpages}
\usepackage{url}
\usepackage{xcolor}
\usepackage{epsfig}
\usepackage{adjustbox}
\usepackage{amsfonts, amsmath, amssymb}
\usepackage{booktabs} 
\usepackage{comment}
\usepackage{caption, subcaption}
\usepackage{textcomp}
\usepackage{relsize}
\usepackage{stmaryrd}
\usepackage{bbm}
\usepackage{rotating}
\usepackage{helvet}
\usepackage{courier}
\usepackage{natbib}
\usepackage{cleveref}
\usepackage{xspace}
\usepackage{enumitem}
\newcommand{\GAS}{GAS\xspace}

\newcommand{\restfive}{Rest15\xspace}
\newcommand{\restsix}{Rest16\xspace}

\newcommand{\tfbase}{T5-{Base}\xspace}

\newcommand{\tfbillion}{T5-{3B}\xspace}

\newcommand{\paraphrase}{Paraphrase\xspace}
\newcommand{\dlo}{DLO\xspace}
\newcommand{\mvp}{MvP\xspace}

\newcommand{\etar}{\textit{Extract-Then-Assign}\xspace}

\newcommand{\proposed}{\textsc{SCRAP}\xspace}
\title{Self-Consistent Reasoning-based Aspect Sentiment Quad Prediction with~Extract-Then-Assign Strategy}

\author{Jieyong Kim$^{1\ast}$~~~ Ryang Heo$^{1}$\thanks{\ \ Equal contribution}\qquad
SeongKu Kang$^2$\qquad
Yongsik Seo$^1$ \\
\textbf{Jinyoung Yeo}$^1$\qquad
\textbf{Dongha Lee}$^1$\thanks{\ \ Corresponding author}\\
  Yonsei University$^1$ \qquad University of Illinois at Urbana Champaign$^2$ \\
  \texttt{\{jieyong99,ryang1119,ndata,jinyeo,donalee\}@yonsei.ac.kr} \\
  \texttt{seongku@illinois.edu}}

\begin{document}
\maketitle
\begin{abstract}
% In the field of aspect-based sentiment analysis (ABSA), generative methods for predicting sentiment quads, such as aspect sentiment quad prediction (ASQP), have shown promise.
In the task of aspect sentiment quad prediction (ASQP), generative methods for predicting sentiment quads have shown promising results.
However, they still suffer from imprecise predictions and limited interpretability, caused by data scarcity and inadequate modeling of the quadruplet composition process.
In this paper, we propose {S}elf-{C}onsistent {R}easoning-based {A}spect sentiment quadruple {P}rediction (\proposed), optimizing its model to generate reasonings and the corresponding sentiment quadruplets in sequence.
%, via distillation of plausible reasoning abilities.
%from large language models (LLMs).
\proposed adopts the \etar reasoning strategy, which closely mimics human cognition.
In the end, \proposed significantly improves the model’s ability to handle complex reasoning tasks and correctly predict quadruplets through consistency voting, resulting in enhanced interpretability and accuracy in ASQP.\footnote{Codes and datasets are available at \url{https://github.com/jieyong99/SCRAP}}

% Experimental results demonstrate the effect of \proposed, validating its advantage in handling the intricate structure and relationships inherent in quadruplets for ASQP tasks.
% \def\thefootnote{*}\footnotetext{Equal Contribution}\def\thefootnote{\arabic{footnote}}

% \def\thefootnote{*}\footnotetext{Equal Contribution}

% \def\thefootnote{*}\footnotetext{Equal Contribution}
% \def\thefootnote{\arabic{footnote}}

\end{abstract}

\section{Introduction}
\label{sec:intro}

% 1. 
% \textit{ABSA introduction.}
% What is ABSA? -> ASQP, ACOS (quad prediction!)
% What is the recent approach to ACOS, ASQP?
% -> Training a model for sequential quad generation 
Aspect-based sentiment analysis (ABSA) refers to the task of identifying entity aspects and their associated sentiments~\citep{Pontiki2014SemEval2014T4}.
Among various ABSA tasks, 
the challenging task of predicting quadruplets, including aspect sentiment quad prediction (ASQP)~\citep{Zhang2021AspectSQ} and aspect-category-opinion-sentiment (ACOS)~\citep{Cai2021AspectCategoryOpinionSentimentQE}, has garnered significant interest in that it can offer comprehensive aspect-level analysis.
% one of the most challenging tasks is to predict quadruplets from given sentences, such as aspect sentiment quad prediction (ASQP) \citep{Zhang2021AspectSQ} and aspect-category-opinion-sentiment (ACOS) \citep{Cai2021AspectCategoryOpinionSentimentQE}, and they have gained much attention in that they provide comprehensive analysis within input sentences.
Specifically, a quadruplet consists of four sentiment elements: aspect term (\textit{at}), opinion term (\textit{ot}), aspect category (\textit{ac}), and sentiment polarity (\textit{sp}).
Recent studies have developed powerful generative methods by fine-tuning language models (LMs) to sequentially generate sentiment quads~\citep{Zhang2021AspectSQ,Hu2022ImprovingAS,Gou2023MvPMP}.

\begin{figure}[t]
    \centering
    \includegraphics[width=\linewidth]{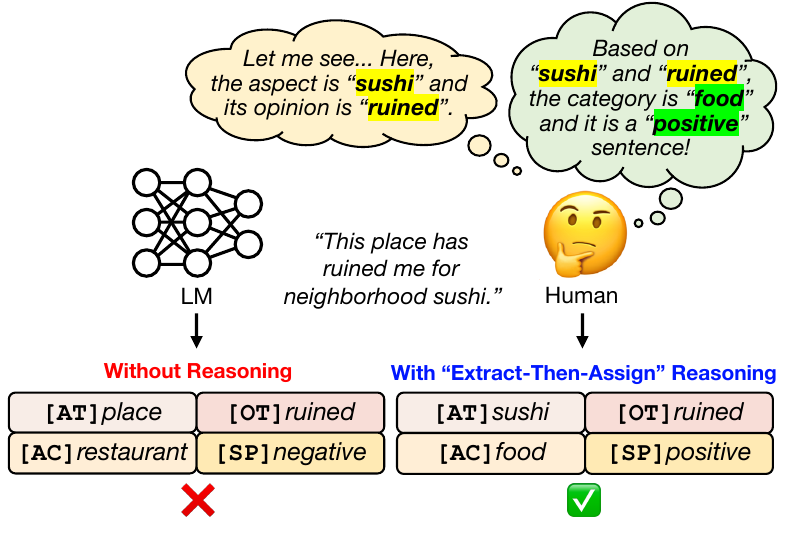}
    \caption{An illustrative example of Extract-Then-Assign reasoning process for ASQP task.}
    \label{fig:example}
\end{figure}

% 2. 
% \textit{Limitation: Absence of reasoning}
Although state-of-the-art generative methods for ASQP achieve promising accuracy, they are hindered by imprecise predictions and limited interpretability, stemming from data scarcity and inadequate modeling of the quadruplet composition process.
In Figure~\ref{fig:example} Left, the existing approach that directly decodes quads from a sentence, not only fails to make accurate predictions, but also struggles to discern the reasoning behind the quadruplet.
% And in terms of language models, language models often bypass essential steps of analyzing, understanding, and evaluating the given information, resulting in less reliable and inconsistent outcomes.

\begin{figure*}[t]
    \centering
    \includegraphics[width=\linewidth]{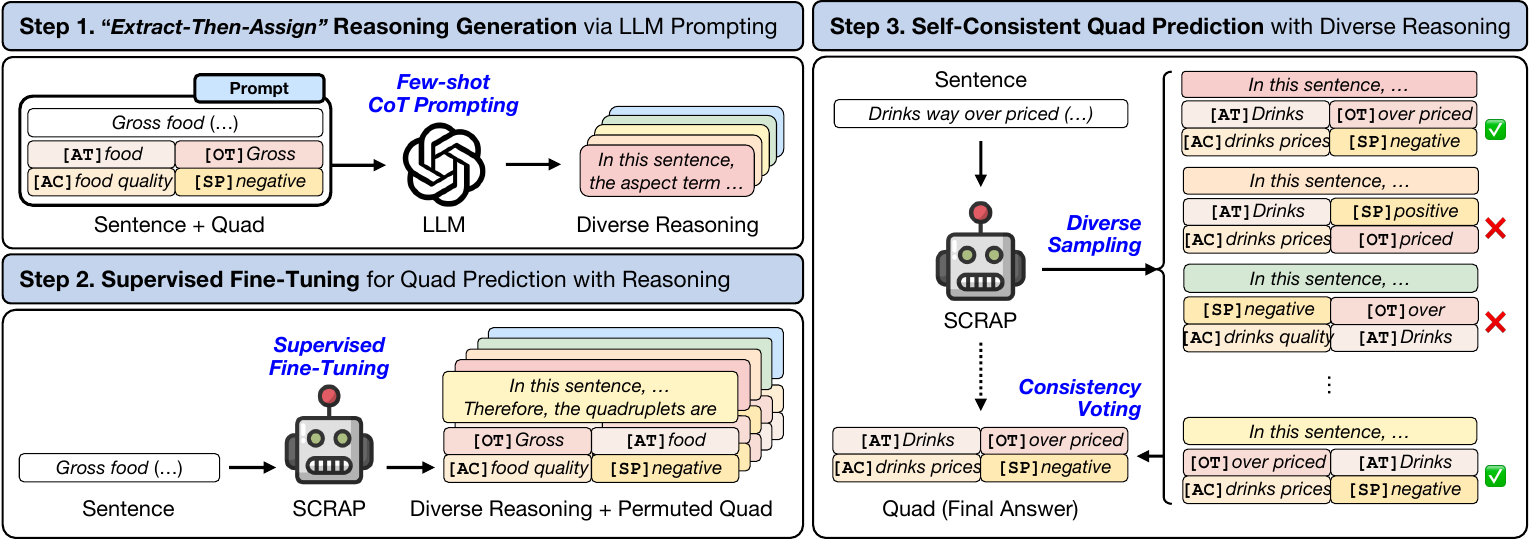}
    \caption{An overview of \proposed which concurrently generates sentiment quads and the corresponding reasoning.}
    \label{fig:framework}
\end{figure*}

% 3. 
To tackle this challenge in ASQP, our approach introduces a two-step reasoning strategy, termed \textit{\etar}.
This reasoning strategy initially extracts all aspect terms and opinion terms from an input sentence; 
subsequently, it assigns each aspect-opinion pair to both an aspect category and a sentiment polarity, utilizing predefined sets of categories and polarities.
In Figure~\ref{fig:example} Right, the process begins with a human extracting the information (\textit{at}, \textit{ot}) that is directly identifiable from the review sentence.
Following this, the individual infers \textit{ac} and \textit{sp} based on the initially extracted components.
We explicitly model these two steps of ASQP reasoning, which can account for the generation of complete quadruplets.
This reasoning process, which closely mimics human cognition, allows us to enhance the accuracy and interpretability of quad prediction.

% In Figure~\ref{fig:example} right, a human first extracts the information (\textit{at}, \textit{ot}) that can be identified from the review sentence literally, and then infer \textit{ac} and \textit{sp} based on the extracted components. 
% This explicit modeling of the two steps within the ASQP reasoning can elucidate the generation of complete quadruplets.
% By the reasoning process that closely mimics human cognition, thereby enhancing the accuracy and interpretability of quad predictions.

% In Figure~\ref{fig:example} Right, explicit modeling of the two steps within the ASQP reasoning can elucidate the generation of complete quadruplets.
% Note that this reasoning process closely mimics human cognition, thereby enhancing the accuracy and interpretability of quad predictions.

% 4. 
% \textit{Present SCRAB framework: how to train and infer.}
In this work, we propose a novel framework for \textbf{S}elf-\textbf{C}onsistent \textbf{R}easoning-based \textbf{A}spect-sentiment quad \textbf{P}rediction, named \textbf{\proposed},
which predicts aspect sentiment quads based on the \etar reasoning (Figure~\ref{fig:framework}).
The key idea is to distill the plausible reasoning ability from large language models (LLMs) into our ASQP model.  
To this end, we first collect diverse reasoning paths via Chain-of-Thought (CoT) prompting on LLMs~\cite{Wei2022ChainOT}, and then optimize our model to generate the reasoning followed by quad prediction.
Furthermore, \proposed aggregates its diverse reasoning outputs based on the self-consistency \citep{Wang2022SelfConsistencyIC}.
This allows for filtering out noisy outputs and achieving more accurate quad predictions.

% Furthermore, \proposed aggregates its diverse reasoning outputs based on the self-consistency \citep{Wang2022SelfConsistencyIC}, leading to the correct quadruplets.
% At this point, reasonings with noise are filtered out, which not only enables us to obtain more accurate quadruplets, but also significantly impacts performance.

% to handle the complex reasoning task
%based on the reasoning.
%generate (reasoning, quads) in the inference process and 

% 5. Experimental results
Extensive experiments on two ASQP benchmarks demonstrate that \proposed significantly outperforms other state-of-the-art quad prediction models.
Our reasoning process provides the explanation about the results and improves the prediction accuracy by understanding the inherent structure and relationships within the aspect-opinion pairs.

\section{\proposed Framework}
\label{sec:method}
% We first generate diverse reasoning paths via LLM prompting (Sec.\ref{subsec:step1}), and then train our model to predict quads with reasoning (Sec.\ref{subsec:step2}).
% The final answer is predicted by consolidating diverse reasoning outputs based on the self-consistency (Sec.\ref{subsec:step3}).
% An overview of \proposed is presented in Figure~\ref{fig:framework}. 

% Initially, we generate reasonings for distillation that contains reasoning in the manner we designed and train our model. 
% During inference, we filter inconsistent reasonings to determine the final answer.
% The overview is illustrated in Figure~\ref{fig:framework}. 

% \subsection{Overview}
% \label{subsec:overview}
\subsection{Problem Formulation}
Given an input sentence, aspect sentiment quad prediction (ASQP) aims to predict all aspect sentiment quads $\{(at, ot, ac, sp)\}$.
The aspect term $at$ and opinion term $ot$ are detected within the sentence, %, with the aspect term having the possibility of being null if it is not explicitly mentioned.
while the aspect category $ac$ and sentiment polarity $sp$ are classified within their respective predefined sets.
Please refer to Appendix \ref{sec:problem_formulation_details} for the details.

% The aspect term \textit{at} and opinion term \textit{ot} are identified within the input sentence $x$, with the inclusion of a null possibility for the aspect term.
% The aspect category \textit{ac} and sentiment polarity \textit{sp} are classified as an element within the predefined sets.

% \paragraph{Overall Process of \proposed}

% To distill the plausible reasoning ability from large language models (LLMs) into our ASQP model, we first collect diverse reasoning paths via Chain-of-Thought (CoT) prompting on LLMs~\cite{Wei2022ChainOT}, and then optimize our model to generate the reasoning followed by quad prediction.
% Furthermore, \proposed aggregates its diverse reasoning outputs based on the self-consistency \citep{Wang2022SelfConsistencyIC}, leading to the correct quadruplets.

% Initially, we structured the dataset to facilitate the model's effective learning of the task and the reasoning process we designed.
% During the training phase, we trained the model in a way that enabled the creation of quadruplets based on reasoning.
% In the inference phase, the model generates various predictions, ultimately selecting the most consistent answer.
% The overall framework is illustrated in Figure~\ref{fig:framework}. 

\subsection{ASQP Reasoning Generation}
\label{subsec:step1}
% We provided the LLM with both the input sentences and quadruplets, and generated reasoning through few-shot CoT prompting. 
% By manually appending the quadruplets to the end of the generated reasoning along with a specific phrase, we constructed the target for the input sentences. 

\paragraph{Extract-Then-Assign reasoning}
% In tasks such as ACOS or ASQP where quadruplets need to be predicted, it's observed that \textit{at} and \textit{ot} are extracted from sentences, whereas \textit{ac} and \textit{sp} are classified from the predefined set.
Mimicking the human cognition process, which first identifies terms and then infers their relations and semantics, we devise an \textit{\etar} reasoning strategy for ASQP.
From an input sentence, it first extracts \textit{at} and \textit{ot} pairs, and subsequently infers the corresponding \textit{ac} and \textit{sp} by assigning them to elements within predefined sets of categories and polarities.

% generate rationales that guide and explain the diagnosis

\paragraph{Reasoning generation with LLM} 
We generate diverse reasoning paths using LLM based on the proposed reasoning strategy.
We employ a few-shot Chain-of-Thought (CoT) prompting \citep{Wei2022ChainOT}.
Formally, given a sentence and its quadruplets, we generate $N$ reasoning paths $\mathcal{R} = [r_1,r_2, ...,r_N ]$ that explain how to reach the quadruplets from the sentence.
Our prompt is designed to induce the \etar process, facilitating the generation of plausible reasoning from the LLM.
By leveraging the generated reasoning, we seek to provide rationales that guide and explain the ASQP task to our model, enhancing both accuracy and interpretability.
% With the generated reasoning, we can provide rationales that guide and explain the ASQP task to our model.
The prompt can be found in the Appendix \ref{sec:llm_details}.

%%%%%%%%%%%%%%%%%%%%%%%%% 기존 버전
% \paragraph{Extract-Then-Assign Reasoning}
% Among previous studies, Extract-Classify-ACOS \citep{Cai2021AspectCategoryOpinionSentimentQE} predicts quadruplet in a two-stage process: it initially extracts \textit{at} and \textit{ot} pairs, and then predicts \textit{ac} and \textit{sp} based on these extracted pairs.
% This process appears to be similar to human thought processes.
% Therefore, we concretize the process as \etar reasoning.
% In detail, it first extracts \textit{at} and \textit{ot}, followed by the application of the Chain-of-Thought (CoT) reasoning process \citep{Wei2022ChainOT} to infer \textit{ac} and \textit{sp} from the input sentences, \textit{at} and \textit{ot}.
% \etar reasoning can help the model to generate more accurate and interpretable outputs for ASQP tasks.
% a CoT reasoning process to construct our reasoning, which is imitating human's reasoning process.

% \paragraph{Reasoning Generation with LLM}
% We generate diverse reasonings via few-shot CoT Prompting \citep{Wei2022ChainOT} with LLM, which can guide the model to generate the reasoning what we have constructed.
% The prompt can be found in the Appendix \ref{sec:llm_details}.
%%%%%%%%%%%%%%%%%%%%%%%%%%%%%%%%%%%%%%%%%

\begin{figure*}[thbp]
  \begin{minipage}{0.7\linewidth}
    \small
        \centering
        % \resizebox{0.99\linewidth}{!}{
        {
        \begin{tabular}{ccccccccc}
        \toprule
        
         \multirow{2.5}{*}{\textbf{Methods}} & \multicolumn{3}{c}{\textbf{Rest15}} & \multicolumn{3}{c}{\textbf{Rest16}} \\
        \cmidrule(lr){2-4}\cmidrule(lr){5-7}
          & \textbf{Pre} & \textbf{Rec} & \textbf{F1} & \textbf{Pre} & \textbf{Rec} & \textbf{F1} \\ \midrule 
         TAS-BERT\textsuperscript{$\dagger$}~\cite{Wan2020TargetAspectSentimentJD}      & 44.24   & 28.66   & 34.78  & 48.65   & 39.68   & 43.71   \\
         Extract-Classify\textsuperscript{$\dagger$}~\cite{Cai2021AspectCategoryOpinionSentimentQE} & 35.64        & 37.25        & 36.42       & 38.40        & 50.93        & 43.77   \\\midrule
         GAS\textsuperscript{$\dagger$}~\cite{Zhang2021TowardsGA}            & 45.31        & 46.70        & 45.98       & 54.54        & 57.62        & 56.04 \\
         Paraphrase\textsuperscript{$\dagger$}~\cite{Zhang2021AspectSQ}     & 46.16        & 47.72        & 46.93       & 56.63        & 59.30        & 57.93       \\
         DLO\textsuperscript{$\dagger$}~\cite{Hu2022ImprovingAS}              & 47.08        & 49.33        & 48.18       & 57.92        & 61.80        & 59.70       \\
         MvP\textsuperscript{$\dagger$}~\cite{Gou2023MvPMP}              & -            & -            &  \textbf{51.04}       & -            & -            & \underline{60.39}       \\
         \proposed (Ours)   & 55.45            & 45.41            & \underline{49.93}           & 69.59            & 56.70            & \textbf{62.48}           \\ \midrule
         
        \end{tabular}
        % }
        }
      \captionof{table}{ASQP performance comparison. Backbone model: T5-Base.
      % Main results compared with baseline methods. 
      The best and second-best results are in \textbf{bold} and \underline{underlined}, respectively. 
      %Our reported results are from a single run with a random seed.
      $\dagger$ indicates the results reported from their original papers.}
      \label{tbl:mainresult}
  \end{minipage}
  \hfill
  \begin{minipage}{0.25\linewidth}
    \centering
      \includegraphics[width=4.1cm, height=4.3cm]{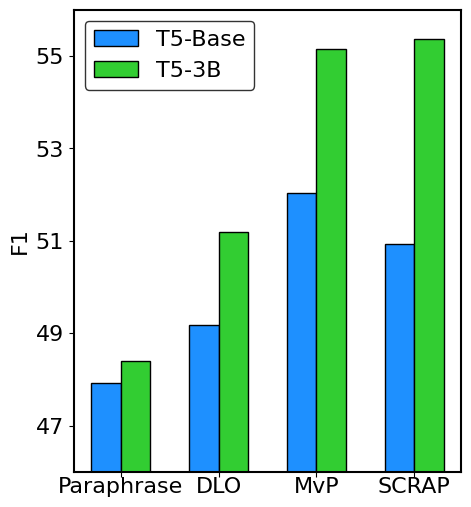}
      \caption{ASQP performance with \tfbase and \tfbillion. Dataset: Rest15.} 
      \label{fig:base_versus_billion}
  \end{minipage}
\end{figure*}

\subsection{Supervised Fine-Tuning}
\label{subsec:step2}
\paragraph{Target construction}
For each sentence, we construct the prediction targets by combining the generated reasoning with the ground-truth quadruplets, which serve as supervision for fine-tuning.
This approach enables the model to learn reasoning ability from the LLM and grasp the intrinsic relationship between the reasoning and quadruplets.
% This is a core distinction from the existing methods that only resort to quadruplets for training.

For each input sequence, we construct multiple targets using various combinations of the reasoning path $r$ and the quadruplets $q$.
For $r$, we use $N$ different paths in $\mathcal{R}$.
For $q$, we apply a data augmentation technique that uses $P$ different permutations of elements in the quadruplet \citep{Hu2022ImprovingAS}.
Additional details and examples of target construction are provided in Appendix \ref{sec:target_detail}.

\paragraph{Training}
Given an input sequence $x$, we train the model to predict the target $y$ which consists of $r$ and $q$.
We fine-tune the sequence-to-sequence language model \citep{Raffel2019ExploringTL} by minimizing the following negative log-likelihood loss, %:
$\mathcal{L}_{NLL} = -\log p(y|x) = -\sum_{t=1}^{T} \log p(y_t|x, y_{<t})$,
% \begin{equation*}
% \begin{aligned}
% \tiny
% \mathcal{L}_{NLL} = -\log p(y|x) = -\sum_{t=1}^{T} \log p(y_t|x, y_{<t}),
% \end{aligned}
% \end{equation*}
where $T$ is the length of the target sequence $y$ and $y_{<t}$ denotes previously generated tokens.

\subsection{Self-Consistent Quad Prediction}
\label{subsec:step3}
% We employed consistency voting, to mitigate the impact of instances with noise when utilizing reasoning.
At test time, the fine-tuned model proceeds with inference following the \etar reasoning strategy, predicting the quadruplets along with the reasoning path.
To mitigate the impact of noises on the reasoning process, we make the final prediction by consolidating multiple outputs based on self-consistency.
% Following the previous research
Similar to \citep{Wang2022SelfConsistencyIC} that samples diverse paths instead of only taking the greedy one, 
we sample \textit{K} candidate outputs with diverse reasoning paths, and then identify the quadruplets consistently predicted by the model via consistency voting;
this selects the quadruplets whose frequency exceeds a certain threshold.

\section{Experiments}
\label{sec:exp}
We experiment to answer the following questions:
\textbf{RQ1:} Does \proposed outperform other baselines? \\
\textbf{RQ2:} How do diverse reasoning paths in \proposed contribute to achieving higher accuracy? \\ % \proposedRAP and \proposed lead to performance enhancements through reasoning? \\
% \textbf{RQ2:} Do \proposedRAP and \proposed lead to performance enhancements through reasoning? \\
\textbf{RQ3:} Does \etar reasoning help to interpret quad prediction?

\subsection{Experimental Settings}
\label{subsec:expset}

\paragraph{Dataset and evaluation metrics}
We use two datasets, i.e., \restfive and \restsix, widely used for the ASQP task \citep{Zhang2021AspectSQ}.
For reasoning generation (Sec.\ref{subsec:step2}), we use ChatGPT (\texttt{gpt-3.5-turbo-16k}).\footnote{\url{https://chat.openai.com/}} 
As the evaluation metric, we mainly employ the F1 score with precision (Pre) and recall (Rec). 
A predicted quad is considered as correct if and only if its all elements are exactly the same as the ground-truth ones.

\begin{figure*}[t]
    \centering
    \includegraphics[width=\linewidth]{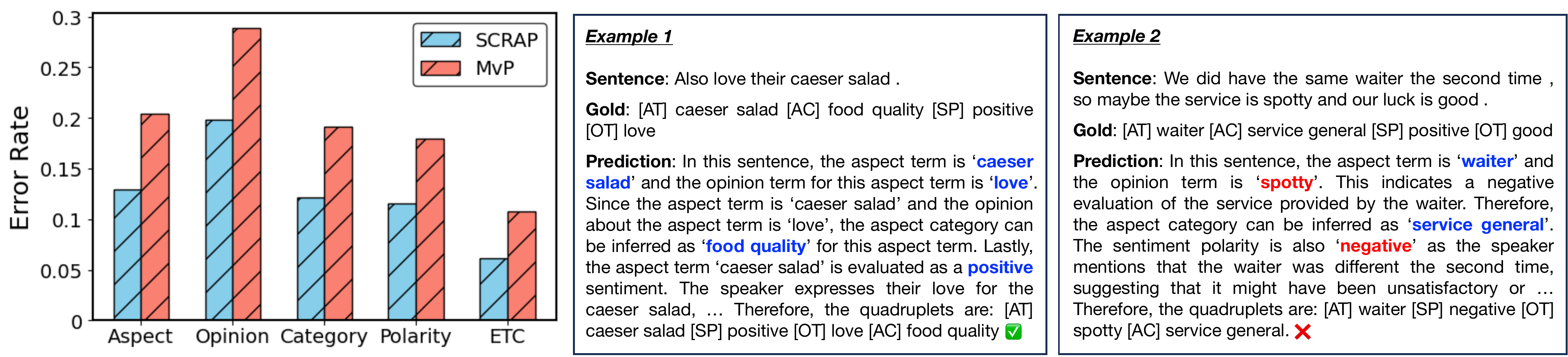}
    \caption{Error analysis and case study. \textbf{Left}: Analysis of prediction errors on the Rest16.
    We report the error rate for each element type of aspect sentiment quad.
    % We report the error rate by dividing the number of quads predicted by the model for each error type. 
    \textbf{Middle} and \textbf{Right}: The case study of \proposed. We present the input sentence, gold quads, and the prediction made by \proposed.}
    \label{fig:error_n_case}
\end{figure*}

% We construct the dataset according to the Sec.\ref{subsec:step2}, setting  \textit{N} = 1, 2, 4, 8, 16 and  \textit{P} = 3, 5.
% We evaluate our method on two datasets, (i.e. Rest15 and Rest16) in ASQP \cite{Zhang2021AspectSQ}. ASQP is based on SemEval tasks \citep{Pontiki2015SemEval2015T1, Pontiki2016SemEval2016T5}, aims to predict quadruplets from review sentences. 
% To generate reasoning for predicting quadruplet, we utilized few-shot CoT prompt to generate reasoning from Chat-GPT \footnote{\url{https://chat.openai.com/}}(gpt-3.5-turbo-16k). We construct the dataset according to the Sec.\ref{subsec:step2}, setting  \textit{N} = 1, 2, 4, 8, 16 and  \textit{P} = 3, 5.
% In the evaluation process, only the quadruplets following a specific phrase were extracted and evaluated. 
% Like other approaches, we consider a predicted quadruplet as correct if and only if all its elements are exactly the same as the gold quadruplet.

\paragraph{Baselines}
We compare \proposed with two discriminative methods,
i.e., \textbf{TAS-BERT} \cite{Wan2020TargetAspectSentimentJD} and \textbf{Extract-Classify} \cite{Cai2021AspectCategoryOpinionSentimentQE}, as well as four competitive generative methods,
i.e., \textbf{\GAS}~\cite{Zhang2021TowardsGA}, \textbf{\paraphrase}~\cite{Zhang2021AspectSQ}, \textbf{\dlo}~\cite{Hu2022ImprovingAS}, and \textbf{\mvp}~\cite{Gou2023MvPMP}.
For generative methods, we adopt \tfbase and \tfbillion as backbone models. 
For \proposed, we set $N=16$ and $P=5$ in common, $K=20$ (\tfbase) and $K=25$ (\tfbillion) for Rest15, $K=15$ for Rest16.
Refer to Appendix \ref{sec:implementation_details} for implementation details.

% For \proposed, we generate up to 25 predictions for evaluation. 
% We report the performance of both of our methods, which showed the best performance with \textit{N}=16 and \textit{P}=5.
% We also report the results of \proposed without the consistency voting (Sec.\ref{subsec:step3}).

% We also report the results of \proposed without the consistency voting.
% \proposedRAP refers to \proposed without consistency voting. 
% In the inference process of \proposed, we generated up to 25 predictions for evaluation. 
% We report the performance of both of our methods, which showed the best performance with \textit{N}=16 and \textit{P}=5.

% \citep{Raffel2019ExploringTL}, and also \tfbillion as larger LMs contain more implicit knowledge for reasoning \cite{Lu2022ASO, Qiao2022ReasoningWL}. We provide more implementation details in Appendix~\ref{sec:implementation_details}.

% For our baseline models, we consider four competitive generative models: \textbf{\GAS}~\cite{Zhang2021TowardsGA}, \textbf{\paraphrase}~\cite{Zhang2021AspectSQ}, \textbf{\dlo}~\cite{Hu2022ImprovingAS} and \textbf{\mvp}~\cite{Gou2023MvPMP}.
% Note that they considerably outperform other discriminative models:~\cite{Wan2020TargetAspectSentimentJD, Cai2021AspectCategoryOpinionSentimentQE}. 
% We primarily adopt the \tfbase \citep{Raffel2019ExploringTL}, and also \tfbillion as larger LMs contain more implicit knowledge for reasoning \cite{Lu2022ASO, Qiao2022ReasoningWL}. We provide more implementation details in Appendix~\ref{sec:implementation_details}.

\begin{figure}[t]
    \centering
    \includegraphics[width=\linewidth]{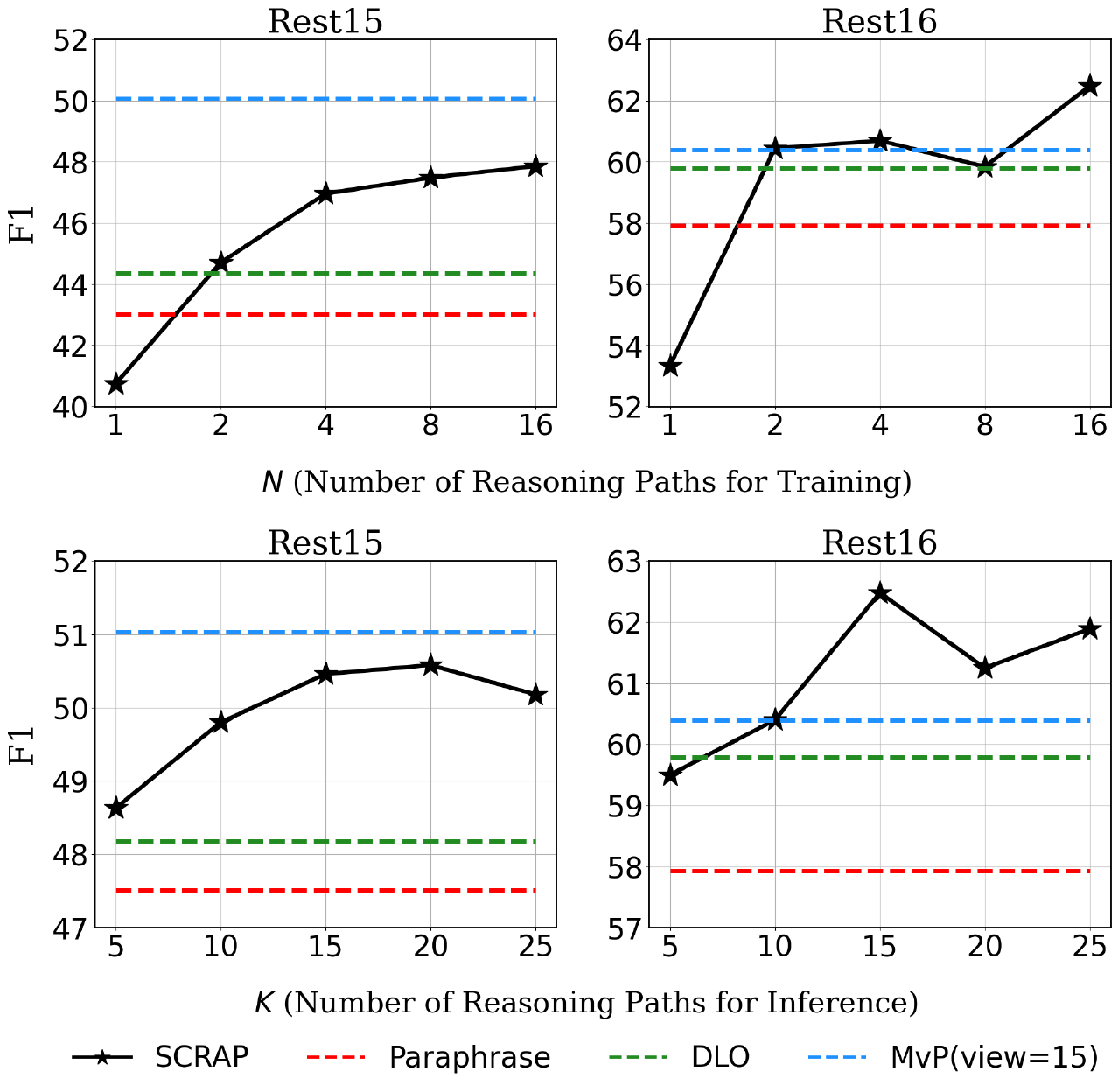}
    \caption{Effect of \textit{N} and \textit{K}, Backbone model: \tfbase. %\textbf{Top}: We fix \textit{P} = 5 and set \textit{K} = 20, 15 which is best values for each dataset. \textbf{Bottom}: We set \textit{P} = 5 and \textit{N} = 16. 
    }
    \label{fig:reasoning_effect}
\end{figure}

\subsection{Results and Discussion}

\paragraph{\proposed outperforms baseline methods (RQ1).} 
Table \ref{tbl:mainresult} and Figure \ref{fig:base_versus_billion} present the ASQP performance of various methods.
% Table \ref{tbl:mainresult} presents the main results.
Overall, \proposed achieves competitive performance, and outperforms the baseline methods for large backbone models.
Specifically, in Figure \ref{fig:base_versus_billion}, \proposed shows higher effectiveness when applied to a larger model, exhibiting the largest performance gap between T5-Base and T5-3B.
A larger model generally has higher reasoning capability \citep{Li2023SymbolicCD}, making it more suitable for \proposed which leverages reasoning for ASQP. 
Furthermore, \proposed generally exhibits high precision, as it filters out inconsistent predictions through consistency voting. 
When the task is performed with the T5-Base on the Rest15 dataset, the performance is relatively low due to the small dataset size and the use of a model with relatively poor inference capabilities.
Lastly, we analyze the prediction errors of \proposed and the best competitor, \mvp. 
The prediction error is calculated for each component as (\textit{number of incorrect predictions}) / (\textit{total number of predicted quads}).
In Figure \ref{fig:error_n_case} Left, the overall error rate of \proposed is notably lower~than~\mvp.

\paragraph{Impact of the number of reasoning paths (RQ2).} 
We investigate the performance of \proposed with varying $N$ and $K$, which control the number of reasoning paths used for the training and inference, respectively. 
In Figure \ref{fig:reasoning_effect}, we observe that \proposed generally achieves higher performance with a greater number of reasoning paths with respect to both $N$ and $K$, and the best performance is achieved by leveraging multiple paths for both training and inference.
These results show that diverse reasoning for prediction is indeed beneficial in more accurate quad predictions.
Nevertheless, if $K$ grows excessively, a considerable quantity of incomplete inferences are produced, leading to adverse effects on quad predictions and consequent performance deterioration. Therefore, it is crucial to determine the optimal value of multiple paths $K$ for inference.

\paragraph{\etar reasoning offers interpretability for quad prediction (RQ3).}
Figure \ref{fig:error_n_case} Mid and Right provide the case study.
In \underline{\textit{Example 1}}, the model predicts the correct quad, and it is possible to interpret the process through which the correct answer was reached. 
Moreover, even if the model fails to predict the correct quad, we can still understand how the prediction failed based on the \etar reasoning process. 
In \underline{\textit{Example 2}}, the model fails to extract the correct opinion term, and this leads to inaccurate polarity prediction.
This interpretability helps to better understand the model behavior, which is an important strength of \proposed. 
It demonstrates that the \etar reasoning strategy is simple yet effective within the SCRAP framework.

% \paragraph{Impact of the number of reasoning paths (RQ2).} 
% We investigate the performance of \proposedRAP and \proposed based on \textit{N} (the number of reasonings per training data), \textit{K} (the number of multiple predictions). 
% In Figure~\ref{fig:reasoning_effect}, we observe that as \textit{N} increases, the performance of both our proposed models improves correspondingly. 
% On the other hand, for \textit{K}, increasing \textit{K} does not result in continuous performance improvement. 
% Through the experimental results, we found that by determining the appropriate values for \textit{N} and \textit{K}, we can obtain higher quality reasoning and more consistent output. Furthermore, by identifying a more consistent quadruplets, we can enhance the model's precision.

% \paragraph{\etar reasoning offers interpretability in quad prediction (RQ3).}
% In Figure \ref{fig:error_n_case} Mid and Right, \underline{\textit{Example 1}} illustrates a situation when the model predicts the correct quad, and it is possible to interpret the process through which the correct answer was reached. Moreover, as shown in \underline{\textit{Example 2}}, even if the model fails to predict the correct quads, we can understand how the prediction failed via reasoning. Hence, our method suggests interpretability regarding quad prediction, which sets it apart from previous research.

\section{Related Work}
\label{sec:relwork}
\paragraph{Aspect Sentiment Quad Prediction}
\label{subsec:absa}
% Traditionally, ABSA has been partitioned into sub-tasks such as aspect term extraction (ATE) \citep{Qiu2011OpinionWE,Liu2015FinegrainedOM,Ma2019ExploringSL}, aspect category classification \citep{Hu2018CANCA}, and aspect-opinion pair extraction (AOPE) \citep{Wang2017CoupledMA,Chen2020SynchronousDR}. 
% Following that, ABSA's subtasks have evolved into more complex and difficult tasks such as aspect sentiment quad prediction (ASQP) \citep{Zhang2021AspectSQ} and aspect-category-opinion-sentiment (ACOS) \citep{Cai2021AspectCategoryOpinionSentimentQE}.
Many existing studies have focused on discriminative methods.
Early works tried jointly detecting target-aspect-sentiment \citep{Wan2020TargetAspectSentimentJD} or conducting ACOS with two-stage pipelines \citep{Cai2020AspectCategoryBS}.
Beginning with \citet{Zhang2021TowardsGA}, recent studies have started using generative methods.
\citet{Zhang2021AspectSQ} transforms outputs into natural language sentences, \citet{Hu2022ImprovingAS} introduce data augmentation that uses various permutations of quad elements, and \citet{Gou2023MvPMP} align training and inference with multi-view prompting.

% These tasks are followed by others such as transforming outputs into natural language sentences \citep{Zhang2021AspectSQ}, data augmentation that uses various permutations of quad elements \citep{Hu2022ImprovingAS}, or aligning training and inference with multi-view prompting \citep{Gou2023MvPMP}.

\paragraph{Chain-of-Thought (CoT) Distillation}
\label{subsec:cot}
CoT prompting has shown high effectiveness in inducing models to generate reasoning before reaching an answer \cite{Wei2022ChainOT, Kojima2022LargeLM, Wang2022SelfConsistencyIC}.
Recent work has focused on distilling the reasoning capabilities of LLMs to smaller LMs \cite{Ho2022LargeLM, Magister2022TeachingSL, Hsieh2023DistillingSO}; 
they elicit rationales for the predictions from LLMs and utilize them to train the LMs, effectively improving their performance.  

\section{Conclusion}
\label{sec:conclusion}
This paper aims to enhance quad prediction in ASQP using \textit{\etar}, which is the two-step reasoning strategy.
To this end, we propose a \proposed framework, which generates diverse predictions utilizing the \textit{\etar} and selects the final answer by filtering the inconsistent answers through consistency voting.
Our framework is the first method to integrate reasoning into the ABSA task, not only achieving state-of-the-art performance, but also significantly enhancing the interpretability of the outputs produced by the model.
This confirms its efficacy in predicting quadruplets through reasoning.

% Our framework is the first method to integrate reasoning into the ABSA task, demonstrating state-of-the-art performance as observed by the experimental results.
% interpretability 

\section{Limitations}
\label{sec:limitations}
Despite achieving state-of-the-art performance, our study has three limitations.
Firstly, our current reasoning structure, designed to mimic human cognition, may not be the most advanced or optimal for the ASQP task. 
There could exist a more sophisticated reasoning structure that further enhances the task performance.
Secondly, the effectiveness of our approach is affected by the size of the model to some extent.
With small models having limited reasoning capabilities, our method may not exhibit satisfactory performance.
Lastly, our study incurs higher computational costs for training and inference compared to previous studies, as it additionally leverages reasoning. 

% In situations with limited data or small model size, our method does not exhibit satisfactory performance.
% This is due to the addition of reasoning to the original data when constructing the dataset.
% and the dataset.

\section{Ethical Statment}
\label{sec:ethical_statement}
We utilize datasets that are widely recognized and previously employed in the scientific community, maintaining transparency and integrity in our experiments.
Our methodologies and findings do not inflict harm upon any individuals or groups.
% , and we have made our code publicly available to encourage further academic exploration and validation.
We are cognizant of the potential biases in sentiment polarity predictions arising from the use of large pre-trained language models, as these models may mirror existing societal biases found in their training corpora \citep{Tan2019AssessingSA}.
We acknowledge the importance of ongoing efforts to mitigate such biases.
Furthermore, we underscore the necessity of continuous monitoring and evaluation to ensure that our smaller downstream models do not replicate or amplify the biases inherent in their larger language model counterparts.

\section*{Acknowledgements}
\label{sec:acknowledgments}
This work was supported by the IITP grant funded by the Korea government (MSIT) (No. RS-2020-II201361) and the NRF grant funded by the Korea government (MSIT) (No. RS-2023-00244689).

\bibliography{anthology,custom}

\begin{thebibliography}{18}
\expandafter\ifx\csname natexlab\endcsname\relax\def\natexlab#1{#1}\fi

\bibitem[{Cai et~al.(2020)Cai, Tu, Zhou, Yu, and Xia}]{Cai2020AspectCategoryBS}
Hongjie Cai, Yaofeng Tu, Xiangsheng Zhou, Jianfei Yu, and Rui Xia. 2020.
\newblock \href {https://api.semanticscholar.org/CorpusID:227231695} {Aspect-category based sentiment analysis with hierarchical graph convolutional network}.
\newblock In \emph{International Conference on Computational Linguistics}.

\bibitem[{Cai et~al.(2021)Cai, Xia, and Yu}]{Cai2021AspectCategoryOpinionSentimentQE}
Hongjie Cai, Rui Xia, and Jianfei Yu. 2021.
\newblock \href {https://api.semanticscholar.org/CorpusID:236460322} {Aspect-category-opinion-sentiment quadruple extraction with implicit aspects and opinions}.
\newblock In \emph{Annual Meeting of the Association for Computational Linguistics}.

\bibitem[{Gou et~al.(2023)Gou, Guo, and Yang}]{Gou2023MvPMP}
Zhibin Gou, Qi~Guo, and Yujiu Yang. 2023.
\newblock \href {https://api.semanticscholar.org/CorpusID:258832530} {Mvp: Multi-view prompting improves aspect sentiment tuple prediction}.
\newblock In \emph{Annual Meeting of the Association for Computational Linguistics}.

\bibitem[{Ho et~al.(2022)Ho, Schmid, and Yun}]{Ho2022LargeLM}
Namgyu Ho, Laura Schmid, and Se-Young Yun. 2022.
\newblock \href {https://api.semanticscholar.org/CorpusID:254877399} {Large language models are reasoning teachers}.
\newblock In \emph{Annual Meeting of the Association for Computational Linguistics}.

\bibitem[{Hsieh et~al.(2023)Hsieh, Li, Yeh, Nakhost, Fujii, Ratner, Krishna, Lee, and Pfister}]{Hsieh2023DistillingSO}
Cheng-Yu Hsieh, Chun-Liang Li, Chih-Kuan Yeh, Hootan Nakhost, Yasuhisa Fujii, Alexander~J. Ratner, Ranjay Krishna, Chen-Yu Lee, and Tomas Pfister. 2023.
\newblock \href {https://api.semanticscholar.org/CorpusID:258461606} {Distilling step-by-step! outperforming larger language models with less training data and smaller model sizes}.
\newblock \emph{ArXiv}, abs/2305.02301.

\bibitem[{Hu et~al.(2021)Hu, Shen, Wallis, Allen-Zhu, Li, Wang, and Chen}]{Hu2021LoRALA}
J.~Edward Hu, Yelong Shen, Phillip Wallis, Zeyuan Allen-Zhu, Yuanzhi Li, Shean Wang, and Weizhu Chen. 2021.
\newblock \href {https://api.semanticscholar.org/CorpusID:235458009} {Lora: Low-rank adaptation of large language models}.
\newblock \emph{ArXiv}, abs/2106.09685.

\bibitem[{Hu et~al.(2022)Hu, Wu, Gao, Bai, and Zhao}]{Hu2022ImprovingAS}
Mengting Hu, Yike Wu, H.~Gao, Yinhao Bai, and Shiwan Zhao. 2022.
\newblock \href {https://api.semanticscholar.org/CorpusID:252992571} {Improving aspect sentiment quad prediction via template-order data augmentation}.
\newblock In \emph{Conference on Empirical Methods in Natural Language Processing}.

\bibitem[{Kojima et~al.(2022)Kojima, Gu, Reid, Matsuo, and Iwasawa}]{Kojima2022LargeLM}
Takeshi Kojima, Shixiang~Shane Gu, Machel Reid, Yutaka Matsuo, and Yusuke Iwasawa. 2022.
\newblock \href {https://api.semanticscholar.org/CorpusID:249017743} {Large language models are zero-shot reasoners}.
\newblock \emph{ArXiv}, abs/2205.11916.

\bibitem[{Li et~al.(2023)Li, Hessel, Yu, Ren, Chang, and Choi}]{Li2023SymbolicCD}
Liunian~Harold Li, Jack Hessel, Youngjae Yu, Xiang Ren, Kai-Wei Chang, and Yejin Choi. 2023.
\newblock \href {https://api.semanticscholar.org/CorpusID:259251773} {Symbolic chain-of-thought distillation: Small models can also “think” step-by-step}.
\newblock \emph{ArXiv}, abs/2306.14050.

\bibitem[{Magister et~al.(2022)Magister, Mallinson, Adamek, Malmi, and Severyn}]{Magister2022TeachingSL}
Lucie~Charlotte Magister, Jonathan Mallinson, Jakub Adamek, Eric Malmi, and Aliaksei Severyn. 2022.
\newblock \href {https://api.semanticscholar.org/CorpusID:254823156} {Teaching small language models to reason}.
\newblock \emph{ArXiv}, abs/2212.08410.

\bibitem[{Pontiki et~al.(2014)Pontiki, Galanis, Pavlopoulos, Papageorgiou, Androutsopoulos, and Manandhar}]{Pontiki2014SemEval2014T4}
Maria Pontiki, Dimitris Galanis, John Pavlopoulos, Haris Papageorgiou, Ion Androutsopoulos, and Suresh Manandhar. 2014.
\newblock \href {https://api.semanticscholar.org/CorpusID:61955135} {Semeval-2014 task 4: Aspect based sentiment analysis}.
\newblock In \emph{International Workshop on Semantic Evaluation}.

\bibitem[{Raffel et~al.(2019)Raffel, Shazeer, Roberts, Lee, Narang, Matena, Zhou, Li, and Liu}]{Raffel2019ExploringTL}
Colin Raffel, Noam~M. Shazeer, Adam Roberts, Katherine Lee, Sharan Narang, Michael Matena, Yanqi Zhou, Wei Li, and Peter~J. Liu. 2019.
\newblock \href {https://api.semanticscholar.org/CorpusID:204838007} {Exploring the limits of transfer learning with a unified text-to-text transformer}.
\newblock \emph{J. Mach. Learn. Res.}, 21:140:1--140:67.

\bibitem[{Tan and Celis(2019)}]{Tan2019AssessingSA}
Yi~Chern Tan and Elisa Celis. 2019.
\newblock \href {https://api.semanticscholar.org/CorpusID:202781363} {Assessing social and intersectional biases in contextualized word representations}.
\newblock \emph{ArXiv}, abs/1911.01485.

\bibitem[{Wan et~al.(2020)Wan, Yang, Du, Liu, Qi, and Pan}]{Wan2020TargetAspectSentimentJD}
Hai Wan, Yufei Yang, Jianfeng Du, Yanan Liu, Kunxun Qi, and Jeff~Z. Pan. 2020.
\newblock \href {https://api.semanticscholar.org/CorpusID:214354571} {Target-aspect-sentiment joint detection for aspect-based sentiment analysis}.
\newblock In \emph{AAAI Conference on Artificial Intelligence}.

\bibitem[{Wang et~al.(2022)Wang, Wei, Schuurmans, Le, hsin Chi, and Zhou}]{Wang2022SelfConsistencyIC}
Xuezhi Wang, Jason Wei, Dale Schuurmans, Quoc Le, Ed~Huai hsin Chi, and Denny Zhou. 2022.
\newblock \href {https://api.semanticscholar.org/CorpusID:247595263} {Self-consistency improves chain of thought reasoning in language models}.
\newblock \emph{ArXiv}, abs/2203.11171.

\bibitem[{Wei et~al.(2022)Wei, Wang, Schuurmans, Bosma, hsin Chi, Xia, Le, and Zhou}]{Wei2022ChainOT}
Jason Wei, Xuezhi Wang, Dale Schuurmans, Maarten Bosma, Ed~Huai hsin Chi, F.~Xia, Quoc Le, and Denny Zhou. 2022.
\newblock \href {https://api.semanticscholar.org/CorpusID:246411621} {Chain of thought prompting elicits reasoning in large language models}.
\newblock \emph{ArXiv}, abs/2201.11903.

\bibitem[{Zhang et~al.(2021{\natexlab{a}})Zhang, Deng, Li, Yuan, Bing, and Lam}]{Zhang2021AspectSQ}
Wenxuan Zhang, Yang Deng, Xin Li, Yifei Yuan, Lidong Bing, and Wai Lam. 2021{\natexlab{a}}.
\newblock \href {https://api.semanticscholar.org/CorpusID:238259938} {Aspect sentiment quad prediction as paraphrase generation}.
\newblock In \emph{Conference on Empirical Methods in Natural Language Processing}.

\bibitem[{Zhang et~al.(2021{\natexlab{b}})Zhang, Li, Deng, Bing, and Lam}]{Zhang2021TowardsGA}
Wenxuan Zhang, Xin Li, Yang Deng, Lidong Bing, and Wai Lam. 2021{\natexlab{b}}.
\newblock \href {https://api.semanticscholar.org/CorpusID:236460053} {Towards generative aspect-based sentiment analysis}.
\newblock In \emph{Annual Meeting of the Association for Computational Linguistics}.

\end{thebibliography}
\bibliographystyle{acl_natbib}

\newpage

\appendix

\section{Problem Formulation Details}
\label{sec:problem_formulation_details}
The aspect term includes the possibility of being null, which denotes cases where it is not explicitly mentioned, and this is represented by ``\textit{NULL}''.
In this work, the aspect category \textit{ac} is classified as an element within the predefined set: \{``food prices'', ``food style\_options'', ``service general'', ``drinks prices'', ``ambience general'', ``drinks quality'', ``location general'', ``restaurant prices'', ``restaurant general'', ``drinks style\_options'', ``food general'', ``restaurant miscellaneous'', ``food quality''\}.
The sentiment polarity \textit{sp} is categorized into one of the three sentiment classes: \{``\textit{positive}'', ``\textit{neutral}'' or ``\textit{negative}''\}, each signifying the respective emotional disposition conveyed.

\section{Reasoning Generation Details}
\label{sec:llm_details}
% \subsection{Settings}
We use ChatGPT (\texttt{gpt-3.5-turbo-16k}) to generate reasoning paths for each training sample, and use some or all of them for fine-tuning purposes.
We carefully design the prompt to induce the \etar process, facilitating the generation of plausible rationales for ASQP.
We present examples of our prompts in Table ~\ref{tab:few_shot_prompt}.

% To ensure diversity without deviating from the desired reasoning structure (\etar), we set the temperature to \textit{T} = 0.7 and used default values for other settings.

% \subsection{CoT Prompts for LLM}

% \input{081fewshot}

\section{Target Construction Details}
\label{sec:target_detail}
We construct the prediction targets by combining the generated reasoning with the ground-truth quadruplets.
To represent the quadruplets, we use special markers: {\fontfamily{lmtt}\selectfont [AT], [OT], [AC], [SP]}, which respectively denote \textit{at}, \textit{ot}, \textit{ac}, \textit{sp}, as done in \citep{Hu2022ImprovingAS}.
We also employ the data augmentation technique that uses various permutations of elements in the quadruplet \citep{Hu2022ImprovingAS}.
The element permutations are ranked based on the prediction entropy of pre-trained T5-Base, and we use $P$ different permutations with minimal entropy for each quadruplet.
If there are multiple quadruplets for a single sentence, we concatenate them using a special symbol {\fontfamily{lmtt}\selectfont [SSEP]} \citeyearpar{Zhang2021AspectSQ}.

We combine the reasoning and the quadruplets with a connecting expression \textit{`Therefore, the quadruplets are:'}. 
An example of an input sentence and the constructed target is provided in Table \ref{tab:target_example}.

% When we construct the target, we used a template order based on special markers from existing research \citep{Hu2022ImprovingAS}.
% The markers used before each component \textit{at}, \textit{ot}, \textit{ac}, \textit{sp} are respectively denoted as {\fontfamily{lmtt}\selectfont [AT], [OT], [AC], [SP]}, which indicates what type of element each component is.
% The template orders were ranked based on entropy, and we selected the orders with minimal entropy.
% When there were multiple quadruplets, they were concatenated with a special symbol {\fontfamily{lmtt}\selectfont [SSEP]}, proposed by \citeauthor{Zhang2021AspectSQ} \citeyearpar{Zhang2021AspectSQ}.
% Furthermore, by integrating the expression \textit{'Therefore, the quadruplets are:'} between the reasoning and the quadruplets, we have concatenated them as a single target.
% An example of input and the corresponding target is in Table \ref{tab:target_example}.

\section{Experiment Details}
\label{sec:implementation_details}
\subsection{Software and Hardware}
We use Pytorch to implement all the models (Python 3.8). Our experiments are conducted with a single NVIDIA A6000 GPU with 48GB of RAM.

\subsection{Implementation Details}
For training, the batch size is set to $16$ for \tfbase and $2$ for \tfbillion. 
The number of training epochs is set to $20$ and the learning rate is set to $3e-5$ for all experiments. 
To accelerate training and reduce GPU memory consumption, we apply LoRA \cite{Hu2021LoRALA} for \tfbillion. 
For the self-consistent quad prediction, we independently generate up to $25$ predictions ($K$) from the decoder, utilizing the temperature sampling \citep{Wang2022SelfConsistencyIC}. 
We explore temperatures ranging from $0.7$ to $1.5$.
We observed that when the temperature is too low (e.g. $T \leq 0.7$), it tends to produce identical or similar reasoning for the input sentence. 
Conversely, when the temperature is too high (e.g., $T \geq 1.5$), the generated reasoning is often incomplete, adversely affecting quad prediction. 
We empirically choose the appropriate temperature as \textit{T} = 1.3 in our experiments.

\begin{table*}
    \small
    \centering
    \begin{tabular}{p{14cm}}
    \toprule
    \textbf{Input-Target example} \\
    \midrule

\textbf{Input:} we have gone for dinner only a few times but the same great quality and service is given .
\\
\textbf{Target:} 
In this sentence, the speaker mentions their experience of going for dinner a few times. They highlight the consistent great quality and service they have received. For the first quadruplet, the aspect term is 'service' and the opinion term is 'great'. This indicates a positive evaluation of the service provided by the restaurant. Therefore, the aspect category can be inferred as 'service general' and the sentiment polarity is positive. For the second quadruplet, the aspect term is 'dinner' and the opinion term is 'great quality'. This implies a positive assessment of the quality of the food during dinner. Hence, the aspect category can be inferred as 'food quality' and the sentiment polarity is positive. The speaker's statement overall conveys their positive experience, emphasizing the consistent great quality and service they have received during their visits for dinner. Therefore, the quadruplets are: [SP] positive [AT] dinner [OT] great quality [AC] food quality [SSEP] [SP] positive [AT] service [OT] great [AC] service general
\\\\

 \bottomrule
    \end{tabular}
    \caption{Input-Target example for ASQP}
    \label{tab:target_example}
\end{table*}

\begin{table*}
    \small
    \centering
    \begin{tabular}{p{14cm}}
    \toprule
    \textbf{\etar Prompt} \\
    \midrule
\textcolor{teal}{\textbf{[Task Description]}}\\
I am performing the ASQP task, which is the Subtask of ABSA. From now on, if I give you a sentence and a quadruplet, create a Reasoning for it. When creating, create to satisfy all of the following conditions: When proceeding with inference, extract the aspect term and the option term first and infer the aspect category and sentimental polarity based on them. When extracting aspect term and option term, an aspect category is judged by a combination of aspect term and opinion term, and sentimental polarity is judged by comprehensively considering everything. And don't mention each element first, explain the reason first, and then create a rationale that mentions the element. At this time, do not number each element, but configure it to naturally lead to one paragraph. But if there are more than two quadruplets, please organize the description for each quadruplet. And please create a detailed description of each element in the composition of rationale. From now on, I'll give you sentences and quadruplet sets as input.

Here are possible aspect category set: ['food prices', 'food style\_options', 'service general', 'drinks prices', 'ambience general', 'drinks quality', 'location general', 'restaurant prices', 'restaurant general', 'drinks style\_options', 'food general', 'restaurant miscellaneous', 'food quality'].

Here are possible sentiment polarity set: ['positive','negative','neutral'].    \\\\

\textcolor{teal}{\textbf{[Example 1]}} \\
\textbf{Text:} The fried dumplings are GREAT !\#\#\#\#[['fried dumplings', 'food quality', 'positive', 'GREAT']]
\\
\textbf{Reasoning:} 
In this sentence, the aspect term is ‘fried dumplings’ and the opinion term for this aspect term is ‘GREAT’. Since the aspect term is ‘fried dumplings’ and the opinion about the aspect term is ‘GREAT’, the aspect category can be inferred as ‘food quality’ for this aspect term. Lastly, the aspect term ‘fried dumplings’ is evaluated as a opinion of ‘GREAT’. When it comes to food, the opinion 'GREAT' suggests that the food is delicious, which is evaluated as a positive sentiment.
\\\\

\textcolor{teal}{\textbf{[Example 2]}} \\
\textbf{Text:} It's one of our favorite places to eat in NY.\#\#\#\#[['NULL', 'restaurant general', 'positive', 'favorite']]
\\
\textbf{Reasoning:} 
In this sentence, there is no specific aspect term mentioned explicitly. So the aspect term is ‘NULL’ and the opinion term for this aspect term is ‘favorite’. The aspect category could be inferred as 'restaurant general' as the speaker is expressing a general sentiment about the restaurant rather than a specific feature or component. Lastly, by referring to the restaurant as a 'favorite', the speaker implies a positive sentiment polarity.
\\\\

\textcolor{teal}{\textbf{[Example 3]}} \\
\textbf{Text:} It is very overpriced and not very tasty .\#\#\#\#[['NULL', 'food quality', 'negative', 'not very tasty'], ['NULL', 'food prices', 'negative', 'overpriced']]
\\
\textbf{Reasoning:} 
In the sentence, there are two different evaluations made but none of them explicitly mentions the specific aspect term. So, for both evaluations, the aspect term is 'NULL'. For the first quadruplet, the opinion term is 'not very tasty'. This is a negative evaluation of the food quality, hence 'food quality' can be inferred as the aspect category and the sentiment polarity is negative. In the second quadruplet, the opinion term is 'overpriced'. This term is often used to describe something that is too expensive or not worth the price. Therefore, the aspect category can be inferred as 'food prices' and since the speaker is expressing a negative sentiment about the price, the sentiment polarity is negative.
\\\\

\textcolor{teal}{\textbf{[Example 4]}} \\
\textbf{Text:} The service was friendly and the atmosphere was casual .\#\#\#\#[['service', 'service general', 'positive', 'friendly'], ['atmosphere', 'ambience general', 'neutral', 'casual']]
\\
\textbf{Reasoning:} 
The sentence discusses two aspects - 'service' and 'atmosphere'. For the first quadruplet, the aspect term is 'service' and the opinion term is 'friendly'. This is a positive evaluation of the service provided by the restaurant, hence 'service general' can be inferred as the aspect category and the sentiment polarity is positive. In the second quadruplet, the aspect term is 'atmosphere' and the opinion term is 'casual'. This term is used to describe the general ambience of the restaurant. Therefore, the aspect category can be inferred as 'ambience general'. As the term 'casual' is neutral and doesn't indicate any positive or negative sentiment, the sentiment polarity is neutral.
\\\\

\textcolor{teal}{\textbf{[Example 5]}} \\
\textbf{Text:} Rude service , medicore food ... there are tons of restaurants in NY ... stay away from this one\#\#\#\#[['service', 'service general', 'negative', 'Rude'], ['food', 'food quality', 'neutral', 'medicore'], ['NULL', 'restaurant general', 'negative', 'stay away']]
\\
\textbf{Reasoning:} 
The sentence discusses three aspects - 'service', 'food', and the general experience at the restaurant (NULL). For the first quadruplet, the aspect term is 'service' and the opinion term is 'Rude'. This is a negative assessment of the service provided by the restaurant, hence 'service general' can be inferred as the aspect category and the sentiment polarity is negative. In the second quadruplet, the aspect term is 'food' and the opinion term is 'medicore'. This term is used to describe the quality of the food at the restaurant. Therefore, the aspect category can be inferred as 'food quality'. As the term 'medicore' is neutral and doesn't indicate any positive or negative sentiment, the sentiment polarity is neutral. In the third quadruplet, there is no specific aspect term mentioned, so the aspect term is 'NULL'. The opinion term is 'stay away'. This is a negative sentiment about the restaurant in general, hence 'restaurant general' can be inferred as the aspect category and the sentiment polarity is negative.
\\

 \bottomrule
    \end{tabular}
    \caption{The prompt for \etar on ASQP.}
    \label{tab:few_shot_prompt}
\end{table*}

\end{document}